\documentclass[11pt,a4paper]{article}
\usepackage[hyperref]{acl2018}
\usepackage{times}
\usepackage{latexsym}
\usepackage{times}
\usepackage{url}
\usepackage{latexsym}
\usepackage{CJKutf8}
\usepackage{booktabs}
\usepackage{amsmath}
\usepackage{amssymb}
\usepackage{enumitem}
\usepackage{avisil}
\usepackage{graphicx}
\usepackage{multirow}
\usepackage{xcolor}
\usepackage{tablefootnote}
\usepackage{stmaryrd}
\usepackage{verbatim}
\usepackage{caption}
\usepackage{subfig}
\usepackage{algorithm}
\usepackage{algorithmic}

\usepackage{url}

\aclfinalcopy 
\def\aclpaperid{975} 

\definecolor{darkgreen}{rgb}{0.0, 0.2, 0.13}

\newcommand{\zh}{Chinese}
\newcommand{\es}{Spanish}
\newcommand{\en}{English}
\newcommand{\zhr}{Zh}
\newcommand{\esr}{Es}
\newcommand{\enr}{En}
\newcommand{\coref}{coreference}
\newcommand{\el}{EL}
\newcommand{\xling}{cross-lingual}
\newcommand{\cm}{{\sc C\&M16}}
\newcommand{\sil}{{\sc Sil18}}

\title{Neural Cross-Lingual Coreference Resolution And Its Application To Entity Linking}

\author{Gourab Kundu \and Avirup Sil \and Radu Florian \and Wael Hamza\\
         IBM Research\\
		 1101 Kitchawan Road\\
	     Yorktown Heights, NY 10598\\
\{gkundu, avi, raduf, whamza\}@us.ibm.com}

\date{}

\begin{document}
\maketitle
\begin{abstract}
We propose an entity-centric neural cross-lingual coreference model that builds on multi-lingual embeddings and language-independent features. We perform both intrinsic and extrinsic evaluations of our model. In the intrinsic evaluation, we show that our model, when trained on English and tested on Chinese and Spanish, achieves competitive results to the models trained directly on Chinese and Spanish respectively. In the extrinsic evaluation, we show that our English model helps achieve superior entity linking accuracy on Chinese and Spanish test sets than the top 2015 TAC system without using any annotated data from Chinese or Spanish.  
\end{abstract}

\section{Introduction}
\label{sec:intro}
Cross-lingual models for NLP tasks are important since they can be used on data from a new language without requiring annotation from the new language \citep{ji2014overview, ji2015overview}. This paper investigates the use of multi-lingual embeddings \citep{faruqui2014improving,upadhyay2016cross} for building cross-lingual models for the task of coreference resolution \citep{ng2002improving,pradhan2012conll}. Consider the following text from a Spanish news article:

\textit{``Tormenta de nieve afecta a 100 millones de personas en \underline{EEUU}. Unos 100 millones de personas enfrentaban el sábado nuevas dificultades tras la enorme tormenta de nieve de hace días en la costa este de \underline{Estados Unidos}.''}

The mentions ``EEUU" (``US" in English) and ``Estados Unidos" (``United States" in English) are coreferent. A coreference model trained on English data is unlikely to coreference these two mentions in Spanish since these mentions did not appear in English data and a regular English style abbreviation of ``Estados Unidos" will be ``EU" instead of ``EEUU". But in the bilingual English-Spanish word embedding space, the word embedding of ``EEUU" sits close to the word embedding of ``US" and the sum of word embeddings of ``Estados Unidos" sit close to the sum of word embeddings of ``United States". Therefore, a coreference model trained using English-Spanish bilingual word embeddings on English data has the potential to make the correct coreference decision between ``EEUU" and ``Estados Unidos" without ever encountering these mentions in training data. 

The contributions of this paper are two-fold. Firstly, we propose an entity-centric neural \xling{} \coref{} model. This model, when trained on \en{} and tested on \zh{} and \es{} from the TAC 2015 Trilingual Entity Discovery and Linking (EDL) Task ~\citep{ji2015overview}, achieves competitive results to models trained directly on \zh{} and \es{} respectively.  Secondly, a pipeline consisting of this \coref{} model and an Entity Linking (henceforth \el{}) model can achieve superior linking accuracy than the official top ranking system in 2015 on \zh{} and \es{} test sets, without using any supervision in \zh{} or \es{}.

Although most of the active \coref{} research is on solving the problem of noun phrase coreference resolution in the Ontonotes data set, invigorated by the 2011 and 2012 CoNLL shared task \citep{pradhan2011conll,pradhan2012conll}, there are many important applications/end tasks where the mentions of interest are not noun phrases. Consider the sentence,

\textit{``(\underline{U.S.} president \underline{Barack Obama} who started ((his) political career) in (\underline{Illinois})), was born in (\underline{Hawaii}).''}

The bracketing represents the Ontonotes style noun phrases and underlines represent the phrases that should be linked to Wikipedia by an \el{} system. Note that mentions like ``U.S." and ``Barack Obama" do not align with any noun phrase. Therefore, in this work, we focus on coreference on mentions that arise in our end task of entity linking and conduct experiments on TAC TriLingual 2015 data sets consisting of English, Chinese and Spanish. 
\section{Coreference Model}
\label{sec:model}
Each mention has a \textit{mention} type (m\_type) of either name or nominal and an \textit{entity} type (e\_type) of Person (PER) / Location (LOC) / GPE / facility (FAC) / organization (ORG) (following standard TAC \citep{ji2015overview} notations).

The objective of our model is to compute a function that can decide whether two partially constructed entities should be coreferenced or not. We gradually merge the mentions in the given document to form entities. Mentions are considered in the order of names and then nominals and within each group, mentions are arranged in the order they appear in the document. Suppose, the sorted order of mentions are $m_1$, $\ldots$, $m_{N_1}$, $m_{N_1+1}$, \ldots, $m_{N_1+N_2}$ where $N_1$ and $N_2$ are respectively the number of the named and nominal mentions. A singleton entity is created from each mention. Let the order of entities be $e_1$, \ldots, $e_{N_1}$, $e_{N_1+1}$, \ldots, $e_{N_1+N_2}$.\\
We merge the named entities with other named entities, then nominal entities with named entities in the same sentence and finally we merge nominal entities across sentences as follows:\\
\textbf{Step 1:} For each \textit{named} entity $e_i$ ($1\leq i \leq N_1$), antecedents are all entities $e_j$ ($1\leq j \leq i-1$) such that $e_j$ and $e_i$ have same e\_type. Training examples are triplets of the form ($e_i$, $e_j$, $y_{ij}$). If $e_i$ and $e_j$ are coreferent (meaning, $y_{ij}$=1), they are merged.\\
\textbf{Step 2:} For each nominal entity $e_i$ ($N_1+1\leq i \leq N_1+N_2$), we consider antecedents $e_j$ such that $e_i$ and $e_j$ have the same e\_type and $e_j$ has some mention that appears in the \textit{same sentence} as some mention in $e_i$. Training examples are generated and entities are merged as in the previous step.\\
\textbf{Step 3:} This is similar to previous step, except $e_i$ and $e_j$ have \textit{no} sentence restriction.\\
\textbf{Features:} For each training triplet ($e_1$, $e_2$, $y_{12}$), the network takes the entity pair ($e_1$, $e_2$) as input and tries to predict $y_{12}$ as output. Since each entity represents a set of mentions, the entity-pair embedding is obtained from the embeddings of mention pairs generated from the cross product of the entity pair. Let $M(e_1,e_2)$ be the set \{($m_i$,$m_j$) $|$ ($m_i$,$m_j$)$\in$ $e_1 \times e_2$\} . For each $(m_i,m_j) \in M(e_1,e_2)$, a feature vector $\phi_{m_i,m_j}$ is computed. Then, every feature in $\phi_{m_i,m_j}$ is embedded as a vector in the real space. Let $v_{m_i,m_j}$ dentote the concatenation of embeddings of all features in $\phi_{m_i,m_j}$. Embeddings of all features except the words are learned in the training process. Word embeddings are pre-trained. $v_{m_i,m_j}$ includes the following language independent features:\\
\textbf{String match:} whether $m_i$ is a substring or exact match of $m_j$ and vice versa (e.g. $m_i$ = ``Barack Obama'' and $m_j$ = ``Obama'')\\
\textbf{Distance:} word distance and sentence distance between $m_i$ and $m_j$ discretized into bins\\
\textbf{m\_type:} concatenation of m\_types for $m_i$ and $m_j$\\
\textbf{e\_type:} concatenation of e\_types for $m_i$ and $m_j$\\
\textbf{Acronym:} whether $m_i$ is an acronym of $m_j$ or vice versa (e.g. $m_i$ = ``United States'' and $m_j$ = ``US'')\\
\textbf{First name mismatch:} whether $m_i$ and $m_j$ belong to e\_type of PERSON with the same last name but different first name (e.g. $m_i$=``Barack Obama'' and $m_j$ = ``Michelle Obama'')\\
\textbf{Speaker detection:} whether $m_i$ and $m_j$ both occur in the context of words indicating speech e.g. ``say'', ``said''\\
In addition, $v_{m_i,m_j}$ includes the average of the word embeddings of $m_i$ and average of the word embeddings of $m_j$. 
\subsection{Network Architecture}
\begin{figure*}
\begin{center}
\includegraphics[width=2\columnwidth]{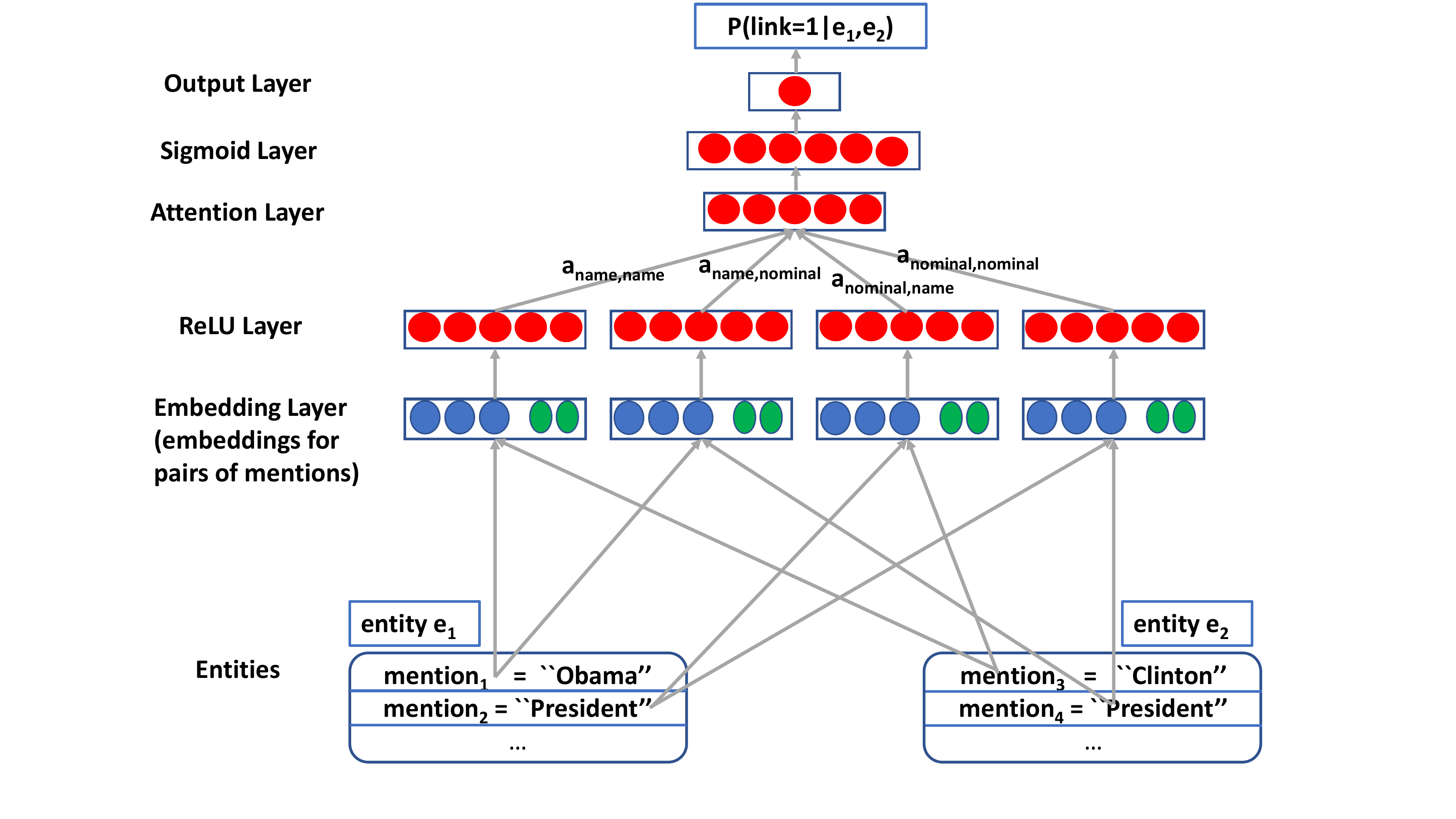}
\end{center}
\caption{Network architecture for our coreference system. Blue circles in mention-pair embeddings layer represent embeddings of features. Green circles represent word embeddings.}
\label{fig:architecture}
\end{figure*}
The network architecture from the input to the output is shown in figure \ref{fig:architecture}.\\
\textbf{Embedding Layer:} For each training triplet ($e_1$, $e_2$, $y$), a sequence of vectors $v_{m_i,m_j}$ (for each ($(m_i, m_j) \in M(e_1,e_2)$)) is given as input to the network.\\
\textbf{Relu Layer:} $v_{m_i,m_j}^r=max(0,W^{(1)}v_{m_i,m_j})$\\
\textbf{Attention Layer:} To generate the entity-pair embedding, we need to combine the embeddings of mention pairs generated from the entity-pair. Consider two entities $e_1$ =\ {(President$^1$, Obama)\} and $e_2$ = \{(President$^2$, Clinton)\}. Here the superscripts are used to indicate two different mentions with the same surface form. Since the named mention pair (Obama, Clinton) has no string overlap, $e_1$ and $e_2$ should not be coreferenced even though the nominal mention pair (President$^1$, President$^2$) has full string overlap. So, while combining the embeddings for the mention pairs, mention pairs with m\_type (name, name) should get higher weight than mention pairs with m\_type (nominal, nominal). The entity pair embedding is the weighted sum of the mention-pair embeddings. We introduce 4 parameters $a_{name, name}$, $a_{name, nominal}$, $a_{nominal, nominal}$ and $a_{nominal, name}$ as weights for mention pair embeddings with m\_types of (name, name), (name, nominal), (nominal, nominal) and (nominal, name) respectively. The entity pair embedding is computed as follows:
\vspace{-1mm}
\begin{multline*}
v_{e_1,e_2}^a =\\
\sum\limits_{(m_i,m_j)\in M(e_1,e_2)}\frac{a_{m\_type(m_i), m\_type(m_j)}}{N}v_{m_i,m_j}^r
\end{multline*}
Here N is a normalizing constant given by:
\vspace{-2mm}
\begin{equation*}
N=\sqrt{\sum\limits_{(m_i,m_j)\in M(e_1,e_2)}a_{m\_type(m_i),m\_type(m_j)}^2}
\end{equation*}
This layer represents attention over the mention pair embeddings where attention weights are based on the m\_types of the mention pairs.\\
\textbf{Sigmoid Layer:} $v_{e_1,e_2}^s = \sigma(W^{(2)}v_{e_1,e_2}^a)
$\\
\textbf{Output Layer:}
\vspace{-3mm}
\begin{equation*}
P(y_{12}=1|e_1,e_2)=\frac{1}{1+e^{-w^s.v_{e_1,e_2}^s}}
\end{equation*}
The training objective is to maximize L.
\vspace{-2mm}
\begin{equation}
L=\prod\limits_{d\in D}\prod\limits_{(e_1,e_2,y_{12})\in S_d} \hspace{-7mm}P(y_{12}|e_1,e_2;W^{(1)},W^{(2)},a,w^s)
\end{equation}
Here $D$ is the corpus and $S_d$ is the training triplets generated from document $d$.

Decoding proceeds similarly to training algorithm, except at each of the three steps, for each entity $e_i$, the highest scoring antecdent $e_j$ is selected and if the score is above a threshold, $e_i$ and $e_j$ are merged.

\section{A Zero-shot Entity Linking model}
We use our recently proposed cross-lingual EL model, described in \cite{sil2018aaai}, where our target is to perform ``zero shot learning'' \cite{socher2013zero,palatucci2009zero}. We train an EL model on English and use it to decode on any other language, provided that we have access to multi-lingual embeddings from English and the target language. We briefly describe our techniques here and direct the interested readers to the paper. The EL model computes several similarity/coherence \textit{scores} $S$ in a ``feature abstraction layer'' which computes several measures of similarity between the context of the mention $m$ in the query document and the context of the candidate link's Wikipedia page which are fed to a feed-forward neural layer which acts as a binary classifier to predict the correct link for $m$. 
Specifically, the feature abstraction layer computes cosine similarities \cite{sil2016one} between the representations of the source query document and the target Wikipedia pages over various granularities. These representations are computed by performing CNNs and LSTMs over the context of the entities. Then these similarities are fed into a Multi-perspective Binning layer which maps each similarity into a higher dimensional vector. We also train fine-grained similarities and dissimilarities between the query and candidate document from multiple perspectives, combined with convolution and tensor networks.

The model achieves state-of-the-art (SOTA) results on English benchmark EL datasets and also performs surprisingly well on Spanish and Chinese. However, although the EL model is ``zero-shot'', the within-document coreference resolution in the system is a language-dependent SOTA coreference system that has won multiple TAC-KBP \cite{ji2015overview,sil2015ibm} evaluations but is trained on the target language. Hence, our aim is to apply our proposed coreference model to the EL system to perform an extrinsic evaluation of our proposed algorithm.
\section{Experiments}
\label{sec:experiments}
We evaluate cross-lingual transfer of coreference models on the TAC 2015 Tri-Lingual EL datasets. It contains mentions annotated with their grounded Freebase \footnote{TAC uses BaseKB, which is a snapshot of Freebase. \sil\ links entities to Wikipedia and in-turn links them to BaseKB.} links (if such links exist) or corpus-wide clustering information for 3 languages: English (henceforth, \enr{}), Chinese (henceforth, \zhr{}) and Spanish (henceforth, \esr{}). Table~\ref{tab:datastats} shows the size of the training and test sets for the three languages. The documents come from two genres of newswire and discussion forums. The mentions in this dataset are either named entities or nominals that belong to five types: PER, ORG, GPE, LOC and FAC.\\
\textbf{Hyperparameters}: Every feature is embedded in a 50 dimensional space except the words which reside in a 300 dimensional space. The Relu and Sigmoid layers have 100 and 500 neurons respectively. We use SGD for optimization with an initial learning rate of 0.05 which is linearly reduced to 0.0001. Our mini batch size is 32 and we train for 50 epochs and keep the best model based on dev set. 
\begin{table}
\small
\begin{center}
\begin{tabular}{cccc}
\toprule
& \enr{} & \esr{} & \zhr{}\\
\midrule
Train & 168 & 129 & 147\\
Test & 167 & 167 & 166\\
\bottomrule
\end{tabular}
\end{center}
\caption{No of documents for the TAC 2015 Tri-Lingual EL Dataset}
\label{tab:datastats}
\end{table}\\
\textbf{Coreference Results:} For each language, we follow the official train-test splits made in the TAC 2015 competition. Except, a small portion of the training set is held out as development set for tuning the models. All experimental results on all languages reported in this paper were obtained on the official test sets. We used the official CoNLL 2012 evaluation script and report MUC, B$^3$ and CEAF scores and their average (CONLL score). See \citet{pradhan2011conll,pradhan2012conll}.

To test the competitiveness of our model with other SOTA models, we train the publicly available system of \citet{clark2016improving} (henceforth, \cm) on the TAC 15 \enr{} training set and test on the TAC 15 \enr{} test set. The \cm\ system normally outputs both noun phrase mentions and their coreference and is trained on Ontonotes. To ensure a fair comparison, we changed the configuration of the system to accept gold mention boundaries both during training and testing. Since the system was unable to deal with partially overlapping mentions, we excluded such mentions in the evaluation. 
Table~\ref{tab:corefeng} shows that our model outperforms \cm\ by 8 points. 

\begin{table}
\small
\begin{center}
\begin{tabular}{ccccc}
\toprule
 &  MUC & B$^{3}$ & CEAF & CoNLL\\
\midrule
This work & 87.8 & 86.8 & 80.9 & \textbf{85.2}\\
\cm & 83.6 & 78.7 & 69.2 & 77.2\\
\bottomrule
\end{tabular}
\end{center}
\caption{Coreference results on the \enr{} test set of TAC 15 competition. Our model significantly outperforms \cm.}
\label{tab:corefeng}
\end{table}

For cross-lingual experiments, we build monolingual embeddings for \enr{}, \zhr{} and \esr{} using the widely used CBOW word2vec model \cite{mikolov2013efficient}. Recently Canonical Correlation Analysis (CCA) \citep{faruqui2014improving}, Multi-CCA \citep{ammar2016massively} and Weighted Regression \citep{mikolov2013exploiting} have been proposed for building the multi-lingual embedding space from monolingual embedding. In our preliminary experiments, the technique of \citet{mikolov2013exploiting} performed the best and so we used it to project the embeddings of \zhr{} and \esr{} onto \enr{}.

In Table~\ref{tab:corefspacmn}, ``\enr{} Model" refers to the model that was trained on the \enr{} training set of TAC 15 using \textit{multi-lingual} embeddings and tested on the \esr{} and \zhr{} testing set of TAC 15. ``\esr{} Model" refers to the model trained on \esr{} training set of TAC 15 using \esr{} embeddings. ``\zhr{} Model" refers to the model trained on the \zhr{} training set of TAC 15 using \zhr{} embeddings. The \enr{} model performs 0.5 point below the \esr{} model on the \esr{} test set. On the \zhr{} test set, the \enr{} model performs only 0.3 point below the \zhr{} model. Hence, we show that without using any target language training data, the \enr{} model with multi-lingual embeddings gives comparable results to models trained on the target language.
\begin{table}
\small
\begin{center}
\begin{tabular}{ccccc}
\toprule
 &  MUC & B$^3$ & CEAF & CoNLL\\
 \hline
\multicolumn{5}{c}{\esr{} Test Set}\\
\midrule
\enr{} model  & 89.5 & 91.2 & 87.2 & 89.3\\
\esr{} Model & 90 & 91.4 & 88 & \textbf{89.8}\\
\midrule
\multicolumn{5}{c}{\zhr{} Test Set}\\
\midrule
\enr{} model & 95.5 & 93.3 & 88.7 & 92.5\\
\zhr{} Model & 96 & 92.8 & 89.6 & \textbf{92.8}\\
\bottomrule
\end{tabular}
\end{center}
\caption{Coreference results on the \esr{} and \zhr{} test sets of TAC 15. \enr{} model performs competitively to the models trained on target language data.}
\label{tab:corefspacmn}
\end{table}\\
\textbf{EL Results:} We replace the in-document coreference system (trained on the target language) of \sil\ with our \enr\ model to investigate the performance of our proposed algorithm on an extrinsic task.
Table \ref{tab:linking} shows the EL results on \esr{} and \zhr{} test sets respectively. ``EL - Coref'' refers to the case where the first step of coreference is not used and EL is used to link the mentions directly to Freebase. ``EL + \enr\ Coref" refers to the case where the neural english coreference model is first used on \zhr{} or \esr{} data followed by the EL model. The former is 3 points below the latter on \esr{} and 2.6 points below \zhr{}, implying coreference is a vital task for EL. Our ``EL + \enr\ Coref'' outperforms the 2015 TAC best system by 0.7 points on \esr{} and 0.8 points on \zhr{}, without requiring any training data for coreference on \esr{} and \zhr{} respectively. Finally, we show the SOTA results on these two data sets recently reported by \sil. Although their EL model does not use any supervision from \esr{} or \zhr{}, their coreference resolution model is trained on a large internal data set on the same language as the test set 
.Without using any in-language training data, our results are competitive to their results (1.2\% below on \esr{} and 0.5\% below on \zhr{}).

\begin{table}
\small
\begin{center}
\begin{tabular}{ccccc}
\toprule
 Systems	& Train on  & Acc. on & Acc. on\\
 & Target Lang &\esr{} &\zhr{}\\
\midrule
EL - Coref & No & 78.1 & 81.3\\
EL + \enr\ Coref & \textbf{No} & \textbf{81.1} & \textbf{83.9}\\
\midrule
TAC Rank 1   & Yes & 80.4 & 83.1\\
\sil & Yes & 82.3 & 84.4\\
\bottomrule
\end{tabular}
\end{center}
\caption{Performance comparison on the TAC 2015 \esr{} and \zhr{} datasets. EL + \enr\ Coref outperforms the best 2015 TAC system (Rank 1) without requiring any \esr{} or \zhr{} coreference data.}
\label{tab:linking}
\end{table}

\section{Related Work}
\label{sec:related}
Rule based \citep{raghunathan2010multi} and statistical coreference models \citep{bengtson2008understanding,rahman2009supervised,fernandes2012latent,durrett2013decentralized,clark2015entity,martschat2015latent,bjorkelund2014learning} are hard to transfer across languages due to their use of lexical features or patterns in the rules. Neural coreference is promising since it allows cross-lingual transfer using multi-lingual embedding. However, most of the recent neural coreference models \citep{wiseman2015learning,wiseman2016learning,clark2015entity,clark2016improving,lee2017end} have focused on training and testing on the same language. In contrast, our model performs cross-lingual coreference. There have been some recent promising results regarding such cross-lingual models for other tasks, most notably mention detection\citep{ni2017weakly} and EL \citep{tsai2016cross,sil2016one}. In this work, we show that such promise exists for coreference also.

The tasks of EL and coreference are intrinsically related, prompting joint models \citep{durrett2014joint,hajishirzi2013joint}. However, the recent SOTA was obtained using pipeline models of coreference and EL \citep{sil2018aaai}. Compared to a joint model, pipeline models are easier to implement, improve and adapt to a new domain.



\section{Conclusion}
\label{sec:conclusion}
The proposed cross-lingual coreference model was found to be empirically strong in both intrinsic and extrinsic evaluations in the context of an entity linking task.

\bibliography{gk}
\bibliographystyle{acl_natbib}

\end{document}